\documentclass[10pt,twocolumn,a4paper]{article}

\usepackage[pagenumbers]{cvpr} 

\usepackage[dvipsnames]{xcolor}

\definecolor{cvprblue}{rgb}{0.21,0.49,0.74}
\usepackage[pagebackref,breaklinks,colorlinks,citecolor=cvprblue]{hyperref}

\usepackage{enumitem}

\usepackage{stfloats}

\usepackage[all]{nowidow}

\usepackage{xcolor}

\usepackage{etoolbox}
\newcounter{BalanceAtReference}
\newcounter{ReferenceIndexForBalancing}
\makeatletter
\def\@balancelastpageonce{%
	\ifnum\value{ReferenceIndexForBalancing}=\value{BalanceAtReference}
	\newpage
	\else
	\relax
	\fi
	\stepcounter{ReferenceIndexForBalancing}
}
\pretocmd{\bibitem}{\@balancelastpageonce}
{} 
{\@latex@error{Patching \bibitem failed}{\@ehd}}
\makeatother

\def\paperID{115} 
\def\confName{3DV\xspace}
\def\confYear{2025\xspace}

\title{VoxelKeypointFusion: Generalizable Multi-View Multi-Person Pose Estimation} 

\author{Daniel Bermuth\\
ISSE\\
University of Augsburg, Germany\\
{\tt\small daniel.bermuth@uni-a.de}
\and
Alexander Poeppel\\
ISSE\\
University of Augsburg\\
{\tt\small poeppel@isse.de}
\and
Wolfgang Reif\\
ISSE\\
University of Augsburg\\
{\tt\small reif@isse.de}
}

\begin{document}
\maketitle

\begin{abstract}
    \vspace{-6pt}
    In the rapidly evolving field of computer vision, the task of accurately estimating the poses of multiple individuals from various viewpoints presents a formidable challenge, especially if the estimations should be reliable as well. This work presents an extensive evaluation of the generalization capabilities of multi-view multi-person pose estimators to unseen datasets and presents a new algorithm with strong performance in this task. It also studies the improvements by additionally using depth information. Since the new approach can not only generalize well to unseen datasets, but also to different keypoints, the first multi-view multi-person whole-body estimator is presented. To support further research on those topics, all of the work is publicly accessible.
\end{abstract}


\vspace{-12pt}
\section{Introduction}
\label{sec:intro}
\vspace{-3pt}

In many applications involving humans, one of the most important tasks is to answer the question: \textit{where are the persons?} In most cases not only the coarse location is of interest, but also the pose of the person, which normally is described by the position of the person's joints. 

There exist several methods to estimate poses from persons, for example by attaching markers to the person's body which can be tracked by special cameras, or by using normal cameras and estimating the pose from their images.
While the marker-based approach is generally more accurate, it is also more complicated because the persons normally need to wear special clothes. This is not convenient and also not always possible, for example, if one likes to track persons in public places, like a shopping store, or if special clothing is required, like in operation rooms. The marker-less approach, on the other hand, is much more convenient, but also more difficult to calculate, since the poses need to be estimated from the images only. 
Usually, multiple cameras are used to capture the scene from different viewpoints, since this makes the estimation more robust to occlusions and can provide more accurate results.

There already exist many works in this field \cite{voxelpose, fastervoxelpose,lin2021multi,dong2019fast,tanke2019iterative}, some of which are presented in the next section. However, most of them are evaluated on the same dataset on which they were trained, and their generalization capabilities were only evaluated on a small scale. 
Therefore, in this work, the focus is put on the generalization capabilities of multi-view multi-person pose estimators, with the target to find a reliable pose estimation with decent speed. For this several state-of-the-art methods, including the one presented in this work, are evaluated on different datasets to analyze their generalization capabilities.

\begin{figure}[htbp]
  \centering
  \begin{subfigure}{0.59\linewidth}
    \centering
    \includegraphics[width=0.99\linewidth]{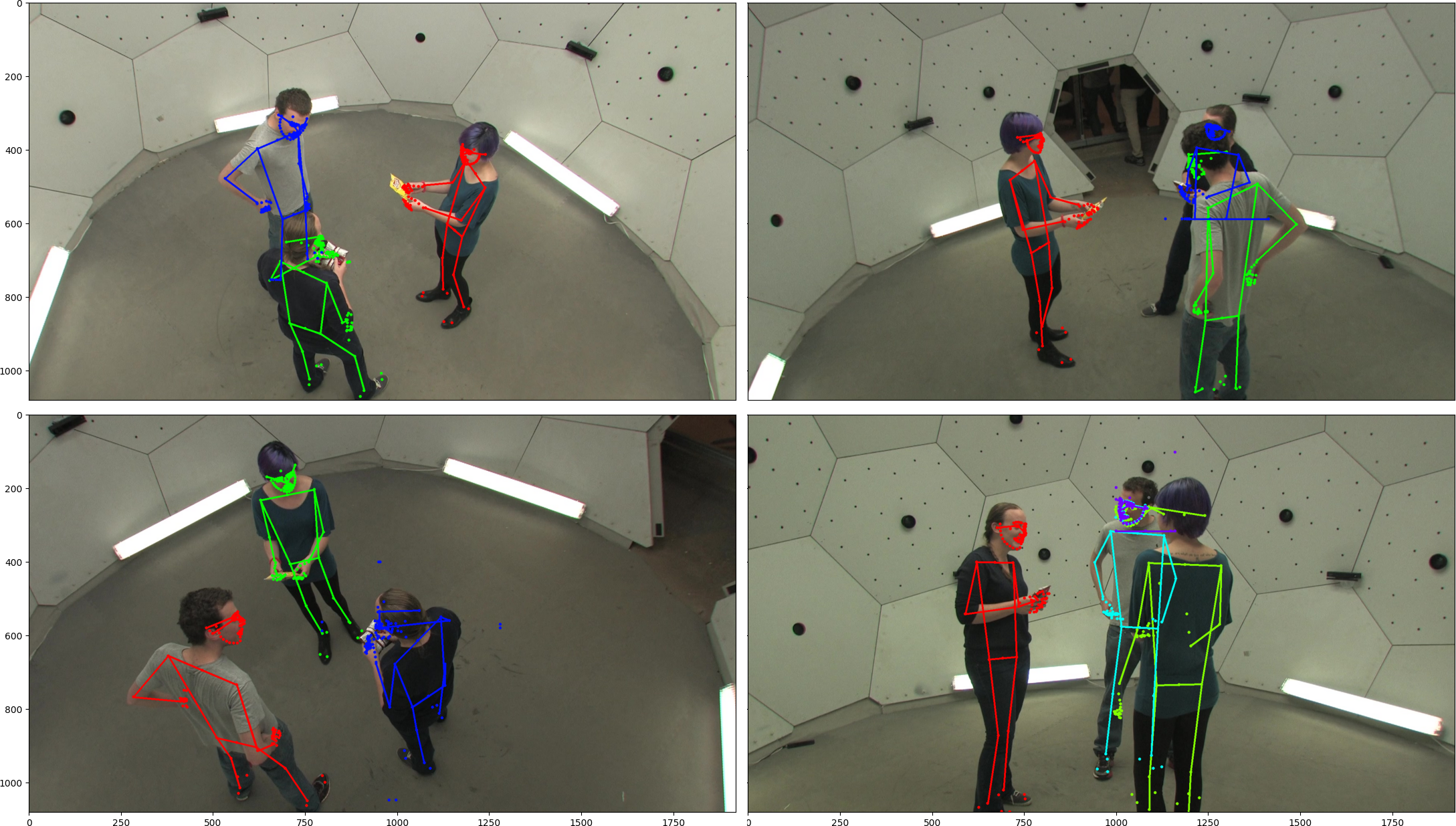}
  \end{subfigure}
  \hfill
  \begin{subfigure}{0.39\linewidth}
    \centering
    \includegraphics[width=0.99\linewidth]{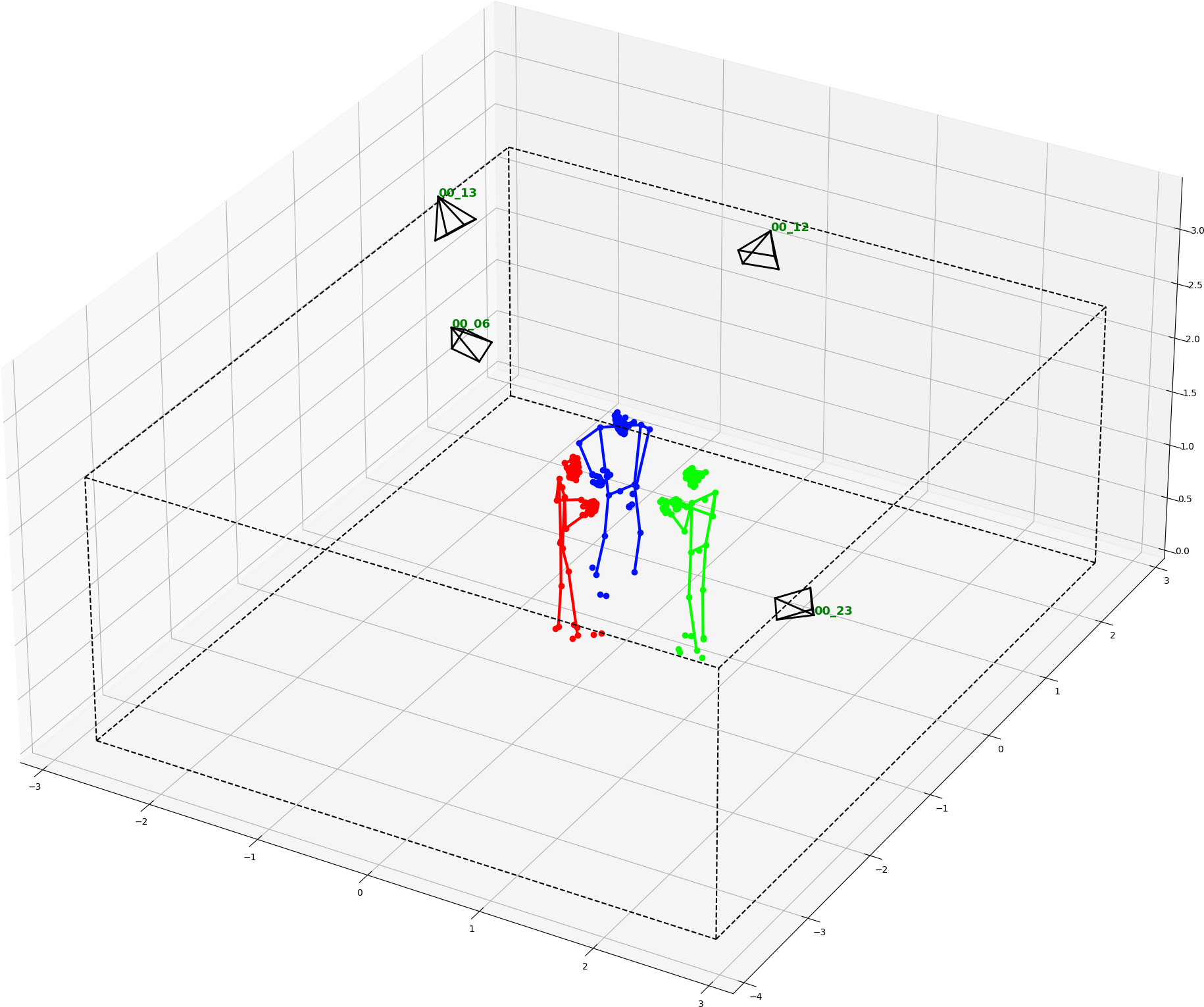}
  \end{subfigure}
  \caption{Example of a multi-person whole-body pose estimation from multiple camera views (from the \textit{panoptic} dataset~\cite{joo2015panoptic}) with \textit{VoxelKeypointFusion}, on the left the per-image 2D estimations, the right shows their fused 3D poses.}
  \label{fig:init_example}
\end{figure}

Currently, almost all approaches only predict the main body joints, like eyes and nose or elbow and wrists, but for some applications, this is not enough. For example, if one likes to analyze the actions of a worker in a human-robot collaboration, the movements of the fingers contain much more information than the movements of the wrists alone. Therefore, it is shown that the presented algorithm can not only generalize well to new datasets, but also to different keypoints, which allows the estimation of whole-body poses~(see Figure~\ref{fig:init_example} for example).

The contributions of this work can be summarized as follows:
(a) the first broad evaluation of the generalization capabilities of multi-view multi-person pose estimators across multiple unseen datasets,
(b) the introduction of a new well generalizing algorithm,
(c) the implementation of the first multi-view multi-person whole-body pose estimator
(d) the introduction of a new library to simplify the handling of dataset related tasks.

The source-code of the presented methods, the evaluation scripts, as well as the dataset handling library, can be found at: \url{https://gitlab.com/Percipiote/}


\section{Related Work I}
\label{sec:relwork}
\vspace{-3pt}

In some cases, it might be possible to reduce the task of estimating human poses to the prediction of a single person only~\cite{iskakov2019learnable, zhang2021adafuse, bartol2022generalizable,chun2023learnable}, which would greatly simplify the problem, because there is no need to match multiple joints to different persons having multiple possible options. However, since it allows a broader range of applications, the focus of this work is put on the multi-person algorithms. It is also assumed that the cameras have been calibrated in advance, a process that should be feasible in the majority of use cases.

The standard approach to solving the estimation problem is a two-step approach, in which the first step is to estimate the 2D poses for each image, and the second step is to estimate the 3D poses from the combined 2D poses. The algorithms can further be divided by whether they solve the 3D estimation problem using an algorithmic approach or a learning-based approach. Another distinction can be made by whether the algorithms make use of temporal information or not. All of them have their own strengths and weaknesses, which can be seen in the later evaluations.

\vspace{3pt}
Regarding the learning-based approaches, \textit{VoxelPose}~\cite{voxelpose} was one of the first concepts, extending the work of \textit{Iskakov et al.}~\cite{iskakov2019learnable} to multi-person estimations. It projects the joint heatmaps from the 2D images into 3D voxelized space and then first estimates a coarse proposal of the persons' centers which are used to generate a more focused 3D space around each person and then calculates the joint locations with a second neural network.
\textit{Faster-VoxelPose}~\cite{fastervoxelpose} is an improved version of \textit{VoxelPose} which uses a more efficient network architecture by reducing the 3D-voxel space to multiple 2D and 1D projections. 
\textit{PRGnet}~\cite{wu2021graph} is a graph-based approach to first detect human centers and then refines them with another graph-network which makes it significantly faster than a plain voxel-based approach. 
\textit{MvP}~\cite{wang2021mvp} directly learns to regress the 3D joint coordinates from the 2D keypoint features of the images without an intermediate volumetric projection.
\textit{PlaneSweepPose}~\cite{lin2021multi} first calculates a cross-view consistency score between the 2D poses of the different views and then regresses the depth in two stages, first a coarse depth estimation for the person and then a fine depth estimation for each joint. The authors published only parts of their code together with preprocessed inputs, which are enough to reproduce the results but not to use the approach on other datasets.
\textit{SelfPose3d}~\cite{Srivastav_2024_CVPR} is a very recent approach based on self-supervised training. It uses the architecture of \textit{VoxelPose}, but only uses predicted 2D poses from another off-the-shelf model which are randomly augmented to train the 2D and 3D networks self-supervised using the additional multi-view information.
\textit{TesseTrack}~\cite{reddy2021tessetrack} also builds upon the concept of \textit{VoxelPose} and adds a time dimension to the voxel space which allows for tracking persons over time and improving poses by using temporal constraints. The source code was not released.
\textit{TEMPO}~\cite{choudhury2023tempo} implemented this in a more efficient approach building upon \textit{Faster-VoxelPose} and tracking bounding boxes instead of adding an additional voxel dimension.  

Regarding the algorithmic approaches, \textit{mvpose}~\cite{dong2019fast} splits the problem into two stages, first finding corresponding 2D poses across the images, by geometric and visual similarity and then triangulating the matched poses. 
\textit{mv3dpose}~\cite{tanke2019iterative} follows a graph matching concept to assign poses using epipolar geometry and also incorporates temporal information to fill in missing joints.
\textit{PartAwarePose}~\cite{chu2021part} uses the poses from the last frame to speed up the matching process and uses a joint-based filter to improve keypoint errors due to occlusions.
\textit{4DAssociation}~\cite{zhang20204d} uses a graph-based concept to match body-parts and connections over space and time to assemble the 3D poses. Evaluating this approach was tried as well, but due to missing code and poor documentation, no way was found to create the required inputs using other datasets.

\vspace{3pt}
In terms of generalization of the learning-based approaches to unseen datasets, a direct transfer is normally not tested, but \textit{VoxelPose}, \textit{Faster-VoxelPose}, \textit{MvP} and \textit{TEMPO} implemented a finetuning concept with synthetic data.
For this, the 3D poses from another dataset are randomly placed in 3D space and back-projected to the camera views of the new dataset. Then those 2D poses, after applying some augmentations, are used to learn the 3D reconstruction again. 
\textit{CloseMoCap}~\cite{shuai2023reconstructing} takes this concept a step further by using a larger 3D pose dataset and implementing additional augmentations to better simulate occlusions and estimation errors. The corresponding source code was not released at the time of writing.
The algorithmic approaches normally evaluate the transfer to a few other datasets, which is much easier since no training is required.


\section{VoxelKeypointFusion}
\label{sec:algs1}
\vspace{-3pt}

The new algorithm called \textit{VoxelKeypointFusion} follows a learning-free algorithmic concept. It can be split into two stages as well, with the first one predicting the 2D poses for each image. For this any 2D pose estimator can be integrated, here \textit{RTMPose}~\cite{jiang2023rtmpose} was used. The second stage can be split into the following steps:

\begin{enumerate}[itemsep=-1pt, topsep=3pt, partopsep=0pt, parsep=3pt, labelindent=9pt]
  \item Generate joint heatmap images for each view
  \item Generate person identity images for each view
  \item Project all heatmaps into a shared voxelized space
  \item Normalize by the number of camera views
  \item Find peaks above a threshold for each joint as keypoint proposals using non-maximum-suppression
  \item Reproject each proposal into all person identity images and get the person-id at this location if existing
  \item Group proposals with the same person ids together
  \item Merge groups with a large overlap of the person ids
  \item Calculate the center of each group and remove outliers
  \item Using the best keypoint proposals, build a person from center to outer limbs
  \item Drop persons with too few keypoints
\end{enumerate}

\begin{figure*}[htbp]
  \centering
  \begin{subfigure}{0.33\linewidth}
    \centering
    \includegraphics[width=0.95\linewidth]{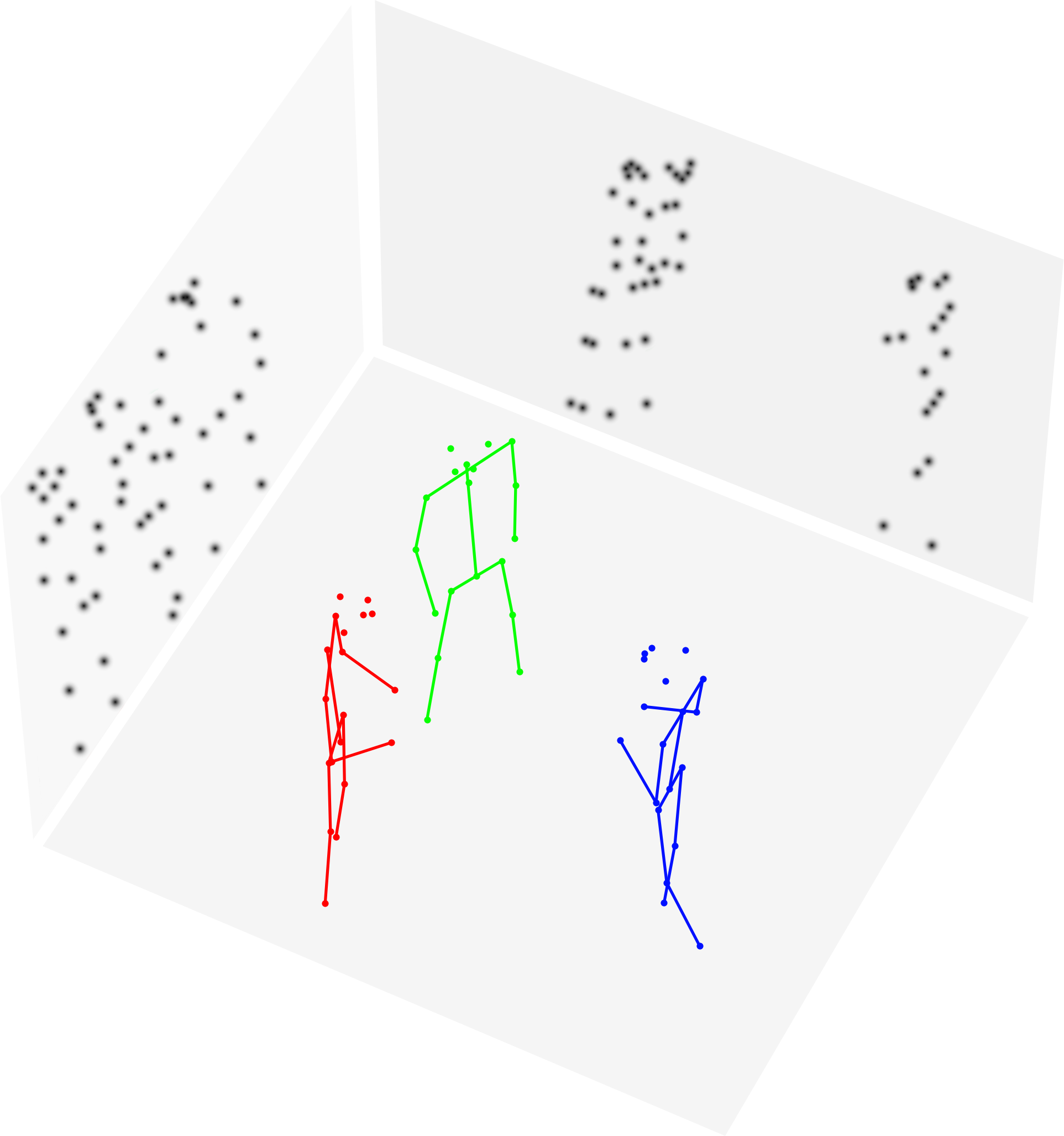}
    \caption{}
    \label{fig:vkf_personids_3}
  \end{subfigure}
  \begin{subfigure}{0.33\linewidth}
    \centering
    \includegraphics[width=0.99\linewidth]{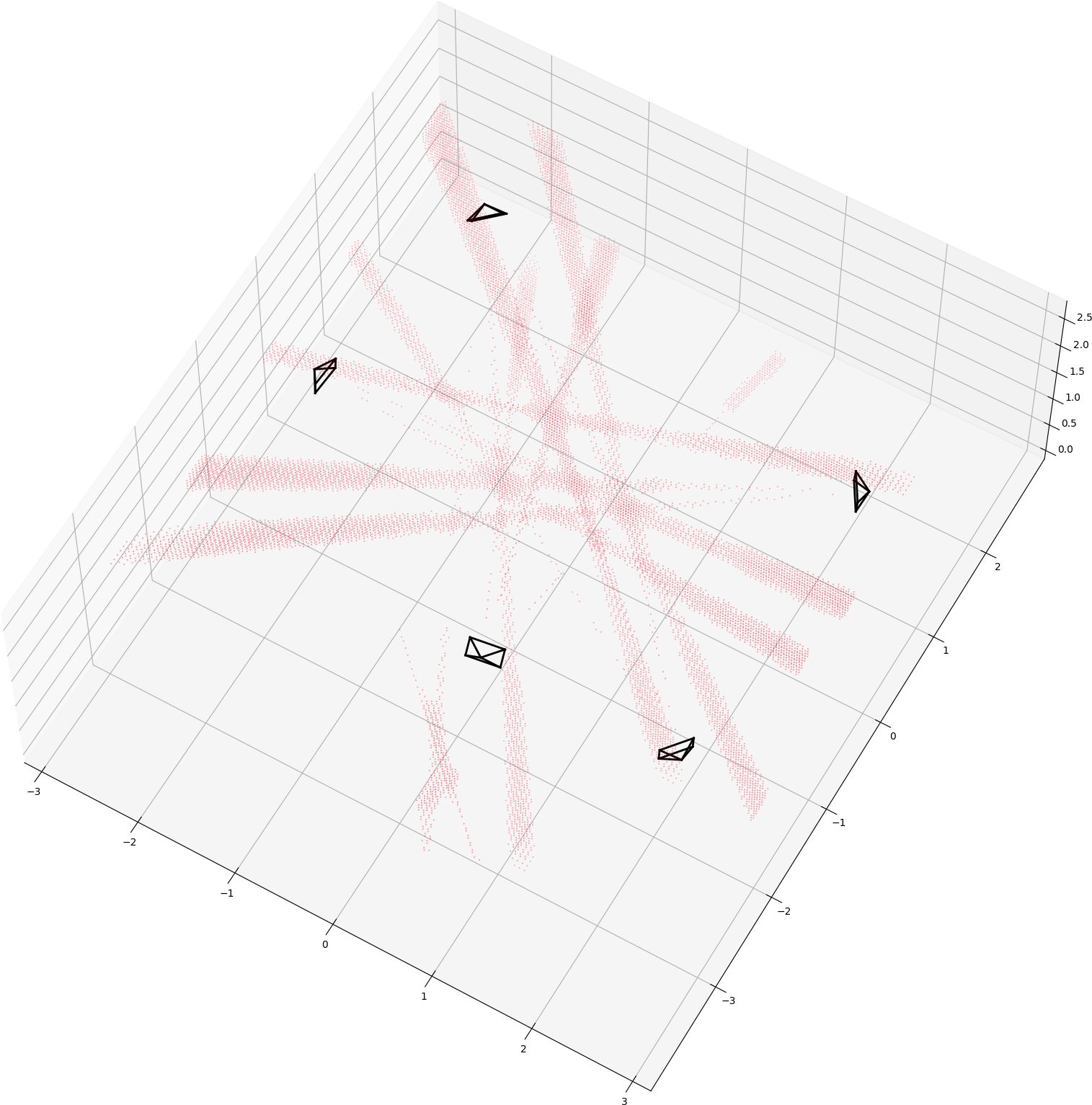}
    \caption{}
    \label{fig:vkf_projection_neck}
  \end{subfigure}
  \hfill
  \begin{subfigure}{0.33\linewidth}
    \centering
    \includegraphics[width=0.95\linewidth]{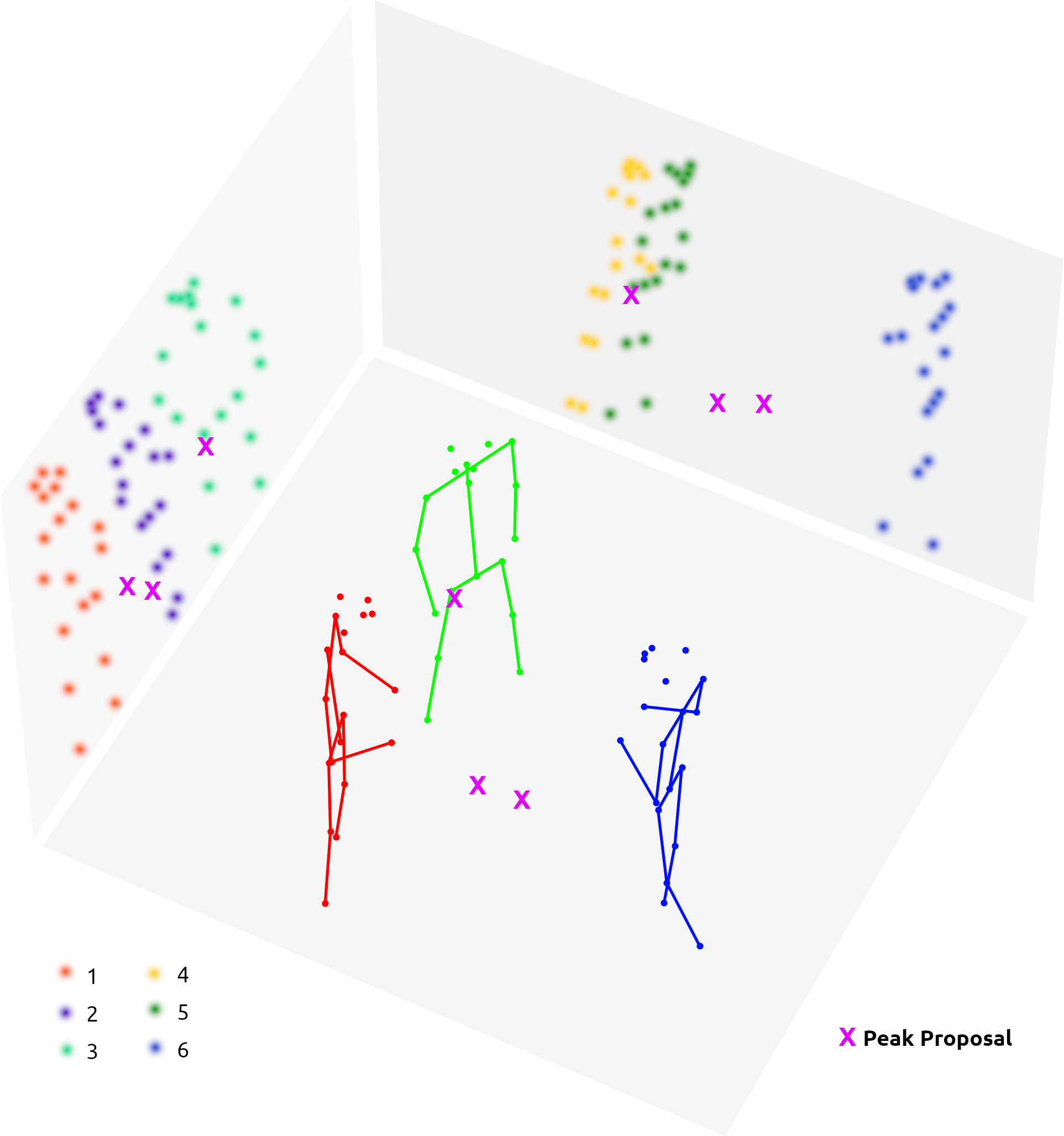}
    \caption{}
    \label{fig:vkf_personids_2}
  \end{subfigure}
  \vspace{-4pt}
  \caption{Obtaining keypoints and person ids. First, the heatmaps of each view (a, schema for two cameras, the 3D-skeletons are drawn only for better visualization) are projected as beams into the voxelized room (b, five real cameras, only a single hip joint visualized). Then peaks at overlapping beam positions are searched and retrieved as keypoint proposals. The peak proposals, here visualized as pink crosses, are projected into each image (c, with the two cameras from a, for better visualization they do not match to overlaps in b or a). The person association is then gathered from the person-id images (c, color-coded and color-paired only for better visualization, so every pixel with a red color has id=$1$). The person-ids of these points are extracted, in this case, one proposal has ids~\{$3$,$5$\}, and the other two received no ids and thus are discarded. The same is done with the proposals from different joint types (so there could be a right shoulder with ids~\{$3$,$5$\}). All proposals with the same id-set are then collected into a person group, which might be merged with other overlapping groups (for example if the left shoulder only got~\{$3$\}).}
  \label{fig:vkf_concept}
\end{figure*}

In step~(1) the keypoints of the 2D pose estimators are used to generate Gaussian heatmap images for each person in the view, which are then combined into a single heatmap. Basically one could also use heatmaps directly from the pose estimator, but not every model has them and it also would need extra implementations for every estimator. In step~(2) images of the same shape are created, but instead of keypoint scores, they contain a person-id everywhere the heatmap score is not zero. The ids are generated by simply counting the total number of persons in all images, and every 2D person is assigned a different number (the order does not matter). Those two steps can be seen on the two cube walls in Figure~\ref{fig:vkf_concept}\,(a+c). The idea is that the information of which joint belongs to which person is already present in the 2D pose estimations. The association of which person is which in the different views is not relevant here, joints of later 3D proposals only need not to be mixed up between different persons in the same view.
\newline
\indent The voxel-based projection (default resolution $50mm$) in steps~(3-5) is quite similar to~\cite{iskakov2019learnable,voxelpose}, except there is no learnable updating/refinement of the scores and only a single voxel-room is used (see Figure~\ref{fig:vkf_concept}, b). The Gaussian heatmap beams overlap at certain points, basically everywhere a joint is seen from at least two views, and sometimes with beams from other persons. The score of each voxel is calculated as the average of the heatmap beams, so overlapping beams lead to higher scores, and well overlapping beams lead to higher scores than poorly overlapping ones. Each voxel over a certain threshold (after non-maximum-neighbor-suppression) is selected as keypoint proposal. The exact location is then adjusted with a weighted average of the neighboring voxels, to reach a sub-voxel precision.
\newline
\indent In step~(6) each keypoint proposal is reprojected to the input views and obtains a set of corresponding person indices from the person-id images if they fall onto a 2D person keypoint (see Figure~\ref{fig:vkf_concept}, c). If the proposal does not match any person, for example when beams from two different persons overlap somewhere in the space, it is discarded directly.
The association to persons then happens in steps~(7+8). In the perfect case, the keypoint proposals of one person all receive the same set of ids and can be clearly matched together. Since this is often not the case, due to occlusions by other persons, which lead to the assignment of wrong ids, groups where the majority of the person ids overlap are merged together, thus ignoring single wrong assignments. 
\newline
\indent In steps~(9-11) one person for each group is created by first taking the best proposal for each keypoint to calculate an initial center, then improving the center by ignoring outliers far away, and then using this second center to drop all keypoints with too large distance in 3D space (above $1.3m$). Then the proposals for the torso joints with the best scores are taken to build an initial person. This person is extended with limbs, by first dropping far away outliers again (above $0.6m$ from the parent joint, since the limb lengths of a person are physically restricted) and assigning the remaining best keypoint proposal. At last any persons with too few keypoints (less than $3$) are dropped as incomplete, and the remaining persons are returned as the final result.

\vspace{3pt}
In a direct comparison with \textit{VoxelPose}, \textit{VoxelKeypointFusion} works without any learnable part. Instead of the top-down approach of first detecting persons, then their joints, a bottom-up approach is used here, which first detects all joints, and then assembles them into persons. This is especially helpful in settings with stronger occlusion, because the risk of totally missing a person is lower.
With the new concept of adding person-ids, already present information about the association of keypoints from the 2D detections is used, which allows skipping the person detection step. 
In difference to previous algebraic/line-based triangulation approaches, the voxel-based approach allows partially overlapping heatmap rays which helps to find a good center between slightly erroneous 2D predictions. The persons are also assembled directly in 3D space, which allows simpler outlier filtering steps.


\section{Skelda}
\label{sec:skelda}
\vspace{-3pt}

Normally most implementations have their own dataset loading tools, which was the same for the evaluated approaches here. But often those datasets are used in multiple implementations and therefore there are many overlapping parts. 
The goal of \textit{skelda} was to extract this into a simple-to-use standalone library, which also contains viewing and evaluation tools. The concept follows some ideas from the audio library \textit{corcua}~\cite{bermuth2021scribosermo} and splits the task into \textit{readers}, \textit{writers} and \textit{evals}. The intermediate format is built from json-like dictionaries, to make it easy to understand the data, and also to simplify adding new datasets as much as possible. 
This also allows merging labels from multiple datasets to create a larger mixed dataset. All datasets used in this work and some others as well can be found in \textit{skelda}, same as the evaluation tools, and all algorithms were extended to use \textit{skelda} to evaluate the different datasets.


\section{Dataset Generalization - Part I}
\label{sec:ndgen1}
\vspace{-3pt}

For the first generalization experiment the \textit{Human36m}~\cite{h36m_pami} dataset is evaluated. The dataset shows a single person moving around and executing different actions and is meanwhile watched by four cameras in the corners of a rectangular room. The algorithms are first directly evaluated on this dataset, using their pretrained models from the \textit{Panoptic}~\cite{joo2015panoptic} dataset, then the learning-based algorithms are retrained with synthetic poses to the new camera positions using their included implementations. The evaluation frames are from the subset $S9$ (commonly $S11$ is used as well, but since it has different camera parameters, one would need to run every synthetic training a second time, which is not very feasible). The images are sampled at a frame rate of $10Hz$ with around $3s$ of consecutive frames and then a larger gap before the next $3s$ of input.

\begin{table*}[hbp]
  \fontsize{7pt}{7pt}\selectfont
  \centering
  \begin{tabular}{@{}|l|c|cc|c|cc|c|c|c|c|@{}}
    \toprule
    Method \hspace{44pt}          & {\,}PCP{\,}   & \multicolumn{2}{c|}{PCK@100/500}          & {\,}MPJPE{\,} & \multicolumn{2}{c|}{Recall@100/500}       & {\,}Invalid{\,} & {}{\,}F1{\,}{} & Training & {\,}FPS{\,}   \\
    \midrule
    VoxelPose                     & 27.1          & \hspace{6pt} 15.6          & 37.3         & 175           & \hspace{6pt} 7.0           & 39.3         & 11.3            & 54.5           & none     & 10.8          \\
    Faster-VoxelPose              & 74.3          & \hspace{6pt} 56.2          & 94.2         & 152           & \hspace{6pt} 58.8          & \textbf{100} & 69.2            & 47.1           & none     & 33.9          \\
    MvP                           & 0             & \hspace{6pt} 0             & 0.1          & 481           & \hspace{6pt} 0             & 0.3          & 99.8            & 0.2            & none     & 9.1           \\
    PRGnet                        & 82.2          & \hspace{6pt} 52.3          & 97.2         & 141           & \hspace{6pt} 27.5          & \textbf{100} & 1.0             & 99.5           & none     & 13.4          \\
    TEMPO                         & 82.0          & \hspace{6pt} 68.6          & 87.5         & 85.5          & \hspace{6pt} 68.5          & 88.3         & 3.3             & 92.3           & none     & 12.4          \\
    SelfPose3d                    & 71.6          & \hspace{6pt} 49.8          & 86.0         & 118           & \hspace{6pt} 38.2          & 87.3         & 77.8            & 35.4           & none     & 7.6           \\
    mvpose                        & 74.5          & \hspace{6pt} 65.0          & 80.6         & 79.3          & \hspace{6pt} 66.0          & 81.3         & 16.6            & 82.4           & none     & 0.3           \\
    mv3dpose                      & 56.1          & \hspace{6pt} 48.1          & 59.5         & 99.8          & \hspace{6pt} 47.3          & 61.0         & \textbf{0}      & 75.8           & none     & 2.7           \\
    PartAwarePose                 & 88.9          & \hspace{6pt} 81.6          & 92.6         & \textbf{64.3} & \hspace{6pt} 83.0          & 93.0         & 0.4             & 96.2           & none     & 4.3           \\
    \midrule
    VoxelPose\,(5\,random)        & 43.1          & \hspace{6pt} 25.7          & 50.1         & 124           & \hspace{6pt} 16.0          & 50.2         & 26.9            & 59.5           & -        & 10.2          \\
    VoxelPose\,(4\,similar)       & 50.4          & \hspace{6pt} 30.7          & 59.0         & 128           & \hspace{6pt} 19.8          & 60.2         & 0.6             & 75.0           & -        & 10.4          \\
    VoxelPose\,(synthetic)        & 90.8          & \hspace{6pt} 74.8          & 98.2         & 92.0          & \hspace{6pt} 70.8          & \textbf{100} & 1.5             & 99.3           & 5h       & 16.0          \\
    Faster-VoxelPose\,(synthetic) & 91.3          & \hspace{6pt} 75.5          & 98.8         & 88.3          & \hspace{6pt} 75.0          & \textbf{100} & 0.2             & 99.5           & 2h       & \textbf{36.2} \\
    \midrule
    \midrule
    VoxelKeypointFusion           & \textbf{96.9} & \hspace{6pt} \textbf{81.7} & \textbf{100} & \textbf{64.3} & \hspace{6pt} \textbf{95.0} & \textbf{100} & \textbf{0}      & \textbf{100}   & none     & 8.7           \\
    \bottomrule
  \end{tabular}
  \caption{Transfer to \textit{human36m} without and with viewpoint training}
  \label{tab:trans_h36m}
\end{table*}

Regarding the metrics, a part in \textit{Percentage of Correct Parts (PCP)} is considered as correct if the average error of two keypoints is lower than the half limb length between those keypoints. Besides arms and legs, the four outer torso connections between hips and shoulder as well as the two shoulders-head/nose connections for the head are scored in all experiments to make them comparable. \textit{Percentage of Correct Keypoints (PCK)} calculates the percentage of how many keypoints were detected with an error lower than the given threshold in millimeters. \textit{Mean Per Joint Prediction Error (MPJPE)} calculates the mean error of all joints for each person and then averages over all persons, dropping persons with an error of above $500mm$ as not matched. In total $13$ keypoints are evaluated (2~shoulders, 2~hips, 2~elbows, 2~wrists, 2~knees, 2~ankles, 1~nose/head). The \textit{Recall} shows the percentage of persons with an average joint error regarding the ground-truth lower than the given threshold. Note that a prediction is always matched to the closest ground-truth person, and in case it gets matched to a second ground-truth person only the better match is kept, whereas the second ground-truth label is counted as not matched. This varies a bit between the evaluations in other works. \textit{Invalid} counts the percentage of predictions that were not matched to any ground-truth label, and \textit{F1} combines it with the \textit{Recall@500} score into a single value. \textit{Training} states the retraining time, if applicable, and \textit{FPS} the end-to-end (images-to-poses) inference speed. Both values were evaluated on a single graphic card, and the \textit{FPS} was measured with a batch-size of~$1$. In most cases a \textit{Nvidia-3090} was used, but for \textit{mvpose}, \textit{mv3dpose} and \textit{PartAwarePose} only a \textit{Nvidia-1080Ti} was working because some of their libraries did not support the newer card.

In Table~\ref{tab:trans_h36m} it can be seen, that \textit{VoxelPose} and especially \textit{MvP} miss most persons if they are directly transferred to the new setup, while \textit{Faster-VoxelPose} and \textit{PRGnet} show much better reliability, but also a quite high joint prediction error. Interestingly the algorithmic approaches have problems finding some of the persons, but achieve a better \textit{MPJPE} than most of the learned algorithms.
Only the newly proposed \textit{VoxelKeypointFusion} shows a reliable performance, while also having a decent prediction speed. In the untrained \textit{MvP} model the number of cameras could not be reduced to four, therefore as a workaround, the first camera was duplicated (in experiments with other datasets this only had a small influence and is not the reason for the poor transfer performance).

Before training \textit{VoxelPose} with synthetic labels and the new camera setup, a transfer with the original \textit{Panoptic}~\cite{joo2015panoptic} dataset was tried. Since this dataset has a large number of cameras, one training with five randomly chosen cameras in each training step, and another one with four fixed cameras with a similar room distribution as in \textit{Human36m}, was executed, but both concepts did not improve the performance much. In difference to that, training with synthetically placed poses and their reprojections to the matching camera views greatly improved the performance.
The synthetic training with \textit{Faster-VoxelPose} significantly improved most metrics as well, especially the number of invalid predictions.
A finetuning of \textit{MvP} and \textit{TEMPO} was not possible due to unpublished tools and missing documentation about how to generate their synthetic data inputs. In their original paper \textit{TEMPO} reported a transfer \textit{MPJPE} of $63.0\,mm$~\cite{choudhury2023tempo}. Reasons for the lower performance here can be differences in the data sampling and the published weights, which already showed a lower performance in the \textit{panoptic} dataset.
\textit{SelfPose3d} followed the concept of \textit{VoxelPose} using data from the \textit{Panoptic} dataset instead of self-supervised data for other datasets than \textit{Panoptic} itself, but those parts were not published as well (the reported results in the paper were slightly worse compared to the ones of \textit{VoxelPose}).

\begin{table*}[hbp]
  \fontsize{7pt}{7pt}\selectfont
  \centering
  \begin{tabular}{@{}|l|c|cc|c|cc|c|c|c|c|@{}}
    \toprule
    Method \hspace{44pt} & {\,}PCP{\,}   & \multicolumn{2}{c|}{PCK@100/500}          & {\,}MPJPE{\,} & \multicolumn{2}{c|}{Recall@100/500}       & {\,}Invalid{\,} & {}{\,}F1{\,}{} & Training & {\,}FPS{\,}   \\
    \midrule
    VoxelPose                     & 90.2          & \hspace{6pt} 78.8          & 91.6         & 63.0          & \hspace{6pt} 88.9          & 91.6         & 57.6            & 58.0           & none     & 7.7           \\
    VoxelPose\,(synthetic)        & 97.7          & \hspace{6pt} 86.3          & 99.4         & 66.7          & \hspace{6pt} 91.0          & \textbf{100} & 48.2            & 68.3           & 5h       & 6.5           \\
    Faster-VoxelPose              & \textbf{99.1} & \hspace{6pt} 88.3          & \textbf{100} & 59.8          & \hspace{6pt} \textbf{99.4} & \textbf{100} & 50.7            & 66.0           & none     & \textbf{17.6} \\
    Faster-VoxelPose\,(synthetic) & 99.0          & \hspace{6pt} 87.9          & \textbf{100} & 58.8          & \hspace{6pt} 99.0          & \textbf{100} & 48.8            & 67.7           & 2h       & 17.4          \\
    MvP                           & 4.6           & \hspace{6pt} 3.2           & 11.0         & 415           & \hspace{6pt} 0             & 15.1         & 95.3            & 7.1            & none     & 8.7           \\
    MvP\,(synthetic)              & 98.6          & \hspace{6pt} \textbf{94.1} & 99.7         & 51.8          & \hspace{6pt} 97.1          & \textbf{100} & 82.2            & 30.2           & -        & 8.5           \\
    PRGnet                        & 98.8          & \hspace{6pt} 86.5          & \textbf{100} & 64.3          & \hspace{6pt} 98.7          & \textbf{100} & 53.3            & 63.7           & none     & 6.3           \\
    TEMPO                         & 96.4          & \hspace{6pt} 86.4          & 97.5         & 57.8          & \hspace{6pt} 95.6          & 97.5         & 50.1            & 66.0           & none     & 5.6           \\
    SelfPose3d                    & 96.3          & \hspace{6pt} 86.1          & 97.5         & 59.9          & \hspace{6pt} 95.8          & 97.5         & 76.1            & 38.4           & none     & 4.9           \\
    mvpose                        & 98.9          & \hspace{6pt} 88.4          & \textbf{100} & 57.3          & \hspace{6pt} 98.1          & \textbf{100} & 51.8            & 65.0           & none     & 0.1           \\
    mv3dpose                      & 97.1          & \hspace{6pt} 91.4          & 98.4         & 55.8          & \hspace{6pt} 94.8          & 98.5         & \textbf{44.3}   & \textbf{71.2}  & none     & 1.6           \\
    PartAwarePose                 & 98.3          & \hspace{6pt} 92.7          & 99.0         & 51.4          & \hspace{6pt} 98.5          & 99.2         & 47.4            & 68.7           & none     & 2.1           \\
    \midrule
    VoxelKeypointFusion           & 98.8          & \hspace{6pt} 93.3          & \textbf{100} & \textbf{51.3} & \hspace{6pt} 98.3          & \textbf{100} & 49.1            & 67.4           & none     & 5.8           \\
    \bottomrule
  \end{tabular}
  \caption{Transfer to \textit{shelf}~\cite{belagiannis20143d}}
  \label{tab:trans_shelf}
\vspace{12pt}
  \fontsize{7pt}{7pt}\selectfont
  \centering
  \begin{tabular}{@{}|l|c|cc|c|cc|c|c|c|c|@{}}
    \toprule
    Method \hspace{44pt} & {\,}PCP{\,}   & \multicolumn{2}{c|}{PCK@100/500}           & {\,}MPJPE{\,} & \multicolumn{2}{c|}{Recall@100/500}       & {\,}Invalid{\,} & {}{\,}F1{\,}{} & Training & {\,}FPS{\,}   \\
    \midrule
    VoxelPose                     & 4.7           & \hspace{6pt} 1.2           & 10.7          & 248           & \hspace{6pt} 0             & 11.2         & 58.8            & 17.6           & none     & 17.9          \\
    VoxelPose\,(synthetic)        & 68.5          & \hspace{6pt} 40.8          & 80.7          & 111           & \hspace{6pt} 26.1          & 80.9         & 27.6            & 76.4           & 6h       & 12.9          \\
    Faster-VoxelPose              & 71.7          & \hspace{6pt} 24.3          & 92.0          & 142           & \hspace{6pt} 4.3           & 92.6         & 48.3            & 66.3           & none     & 23.9          \\
    Faster-VoxelPose\,(synthetic) & 65.9          & \hspace{6pt} 41.3          & 74.4          & 104           & \hspace{6pt} 39.6          & 74.7         & \textbf{2.1}    & 84.8           & 2h       & \textbf{25.3} \\
    MvP                           & 2.4           & \hspace{6pt} 1.2           & 5.7           & 301           & \hspace{6pt} 0             & 6.9          & 95.3            & 5.6            & none     & 8.9           \\
    MvP\,(synthetic)              & 79.3          & \hspace{6pt} 64.0          & 89.3          & 113           & \hspace{6pt} 63.6          & 92.0         & 80.3            & 32.4           & 2h       & 8.9           \\
    PRGnet                        & 2.4           & \hspace{6pt} 0.3           & 8.2           & 333           & \hspace{6pt} 0             & 9.6          & 53.2            & 15.9           & none     & 16.8          \\
    TEMPO                         & 47.7          & \hspace{6pt} 29.8          & 54.4          & 106           & \hspace{6pt} 22.6          & 54.5         & 51.5            & 51.3           & none     & 7.3           \\
    SelfPose3d                    & 24.7          & \hspace{6pt} 6.7           & 36.9          & 207           & \hspace{6pt} 0             & 39.1         & 88.2            & 18.2           & none     & 6.8           \\
    mvpose                        & 91.3          & \hspace{6pt} 70.2          & \textbf{99.9} & 80.4          & \hspace{6pt} 82.7          & \textbf{100} & 25.4            & 85.5           & none     & 0.5           \\
    mv3dpose                      & 84.1          & \hspace{6pt} 64.4          & 93.4          & 135           & \hspace{6pt} 62.5          & 94.9         & 10.1            & \textbf{92.4}  & none     & 2.8           \\
    PartAwarePose                 & \textbf{93.2} & \hspace{6pt} \textbf{78.4} & 98.9          & \textbf{74.7} & \hspace{6pt} \textbf{93.9} & 98.9         & 22.7            & 86.8           & none     & 5.8           \\
    \midrule
    VoxelKeypointFusion           & 91.1          & \hspace{6pt} 74.6          & \textbf{99.9} & 84.4          & \hspace{6pt} 80.3          & \textbf{100} & 35.5            & 78.4           & none     & 7.8           \\
    \bottomrule
  \end{tabular}
  \caption{Transfer to \textit{campus}~\cite{belagiannis20143d} }
  \label{tab:trans_campus}
\end{table*}

\vspace{3pt}
The second evaluation was run with the \textit{Shelf\,\&\,Campus}~\cite{belagiannis20143d} datasets. \textit{Shelf} shows multiple persons assembling a shelf, from five different viewpoints, while \textit{Campus} shows an outside scene with people walking and standing on a place in front of a building, watched by three cameras. Both datasets are commonly used for generalization tests and were also used in most models evaluated here. In difference to most other evaluations that only report \textit{PCP}, more metrics are used, and the \textit{PCP} score is calculated as described above, and not by first averaging over the persons and then merging them together like usually, because this, on the one hand, results in a different weighting of errors since some persons occur more often than others, and on the other hand, allows a better comparison with other datasets.

In Table~\ref{tab:trans_shelf} it can be seen that the learnable approaches show similar performance on the \textit{Shelf} dataset compared to the algorithmic ones. They are notably faster than most others, which mainly is caused by faster 2D pose estimators, but on the other hand, tend to predict more invalid persons. Note that in the dataset itself not all occurring persons are labeled, and therefore the percentage can not be zero, but the algorithmic models can detect all labeled persons while predicting a lower total number of persons. \textit{MvP} shows an especially poor performance in this regard, even with the synthetically finetuned model, making it basically unusable for real-world applications (the preprocessed data and tools for \textit{Shelf\,\&\,Campus} were published so it could be trained here).
While \textit{VoxelPose} profits from the synthetic training, the difference for \textit{Faster-VoxelPose} is only marginal or even slightly worse in some metrics. The camera setup of \textit{Shelf} is quite similar to the \textit{Panoptic} dataset, which is a reason the directly transferred models show a decent performance. In all cases, the models were evaluated end-to-end, instead of using the provided external heatmaps created by unpublished tools, since the generalization of the complete model is the interesting part here.

On the \textit{Campus} dataset, in Table~\ref{tab:trans_campus}, most learned algorithms predict even more invalid persons. \textit{PRGnet} seems not to generalize to this setup very well, but implementing a similar finetuning concept might be able to help here. \textit{VoxelKeypointFusion} also has a problem with predicting too many invalid persons, which is mostly caused by persons standing directly in front of each other from the camera perspective, thus leading to person groups that are not merged well and resulting in more person outputs. This seems to be a general problem of the voxel-based algorithms.
While all models improved their performance through synthetic training, they still do not detect all persons, and generally less than the algorithmic ones, which indicates some problems in their person-proposal algorithms.
It is also complicated to declare a single best approach here since all algorithms have strengths and weaknesses.


\section{Using Depth Information}
\label{sec:algs2}
\vspace{-3pt}

All previously evaluated algorithms only use RGB images to estimate the poses. Since some cameras can also record depth data besides the color images, this information could be helpful to improve the predictions.

\begin{figure*}[!hbp]
  \centering
  \begin{subfigure}{0.33\linewidth}
    \centering
    \includegraphics[width=0.9\linewidth]{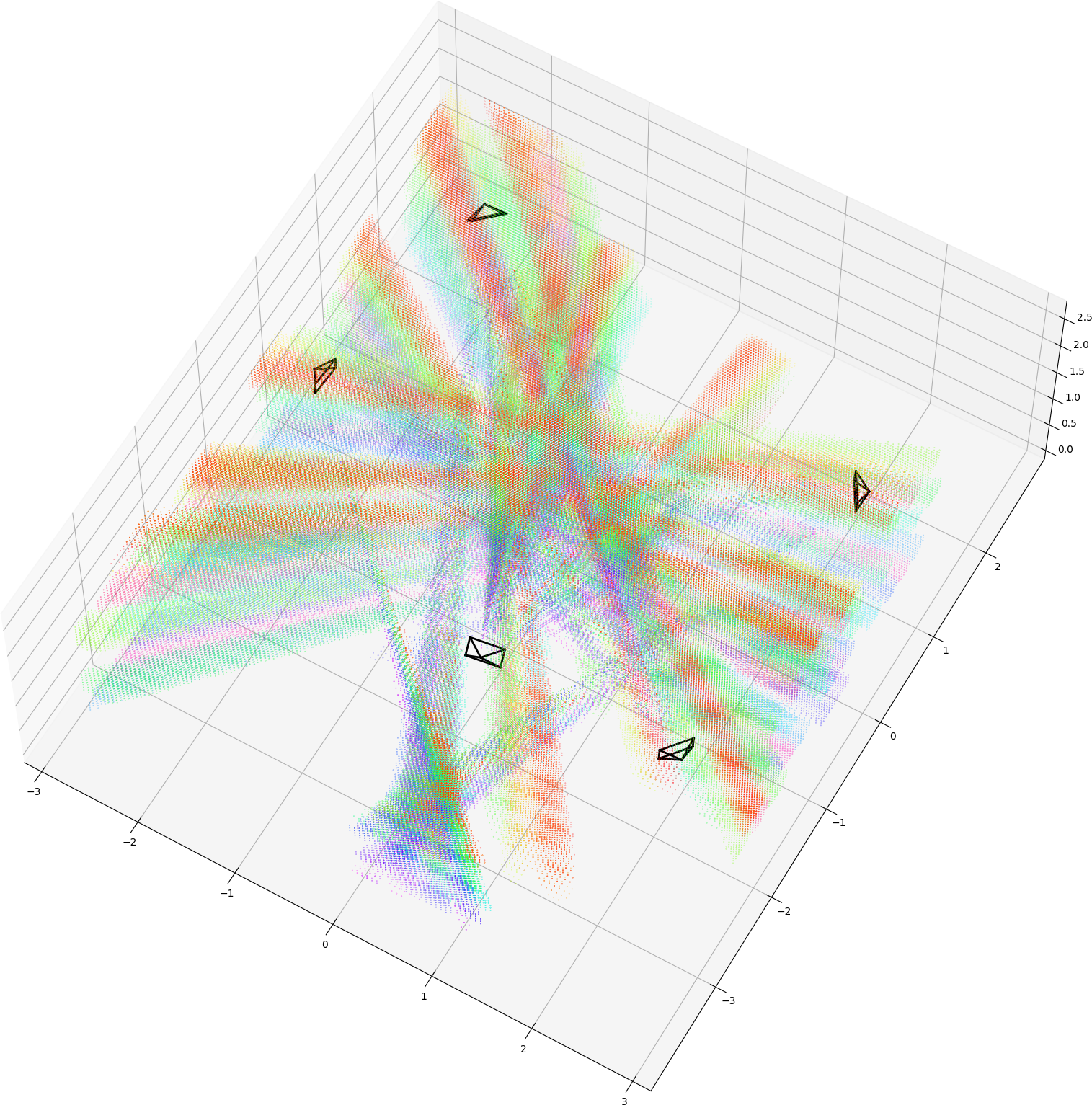}
  \end{subfigure}
  \begin{subfigure}{0.33\linewidth}
    \centering
    \includegraphics[width=0.9\linewidth]{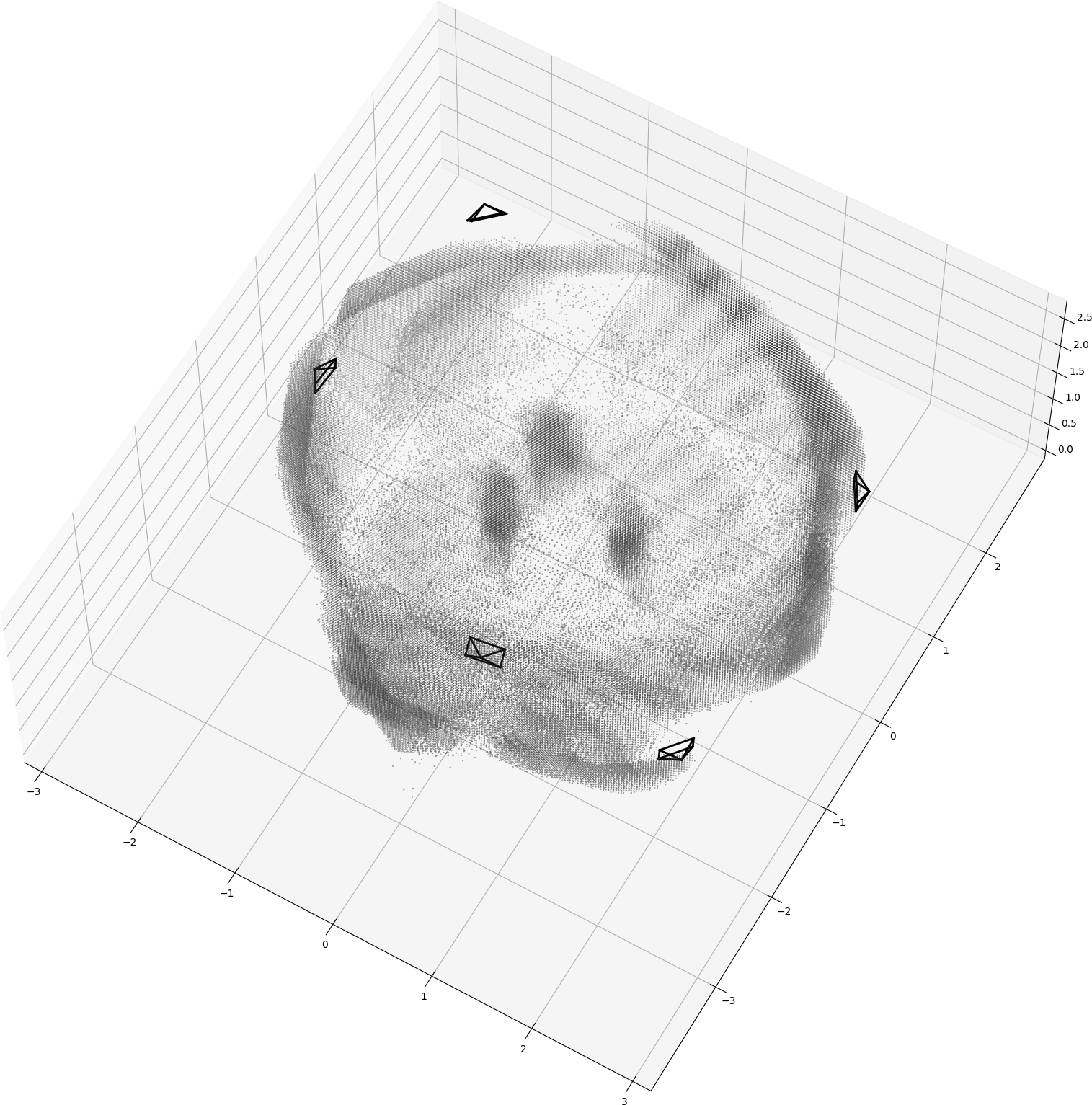}
  \end{subfigure}
  \begin{subfigure}{0.33\linewidth}
    \centering
    \includegraphics[width=0.9\linewidth]{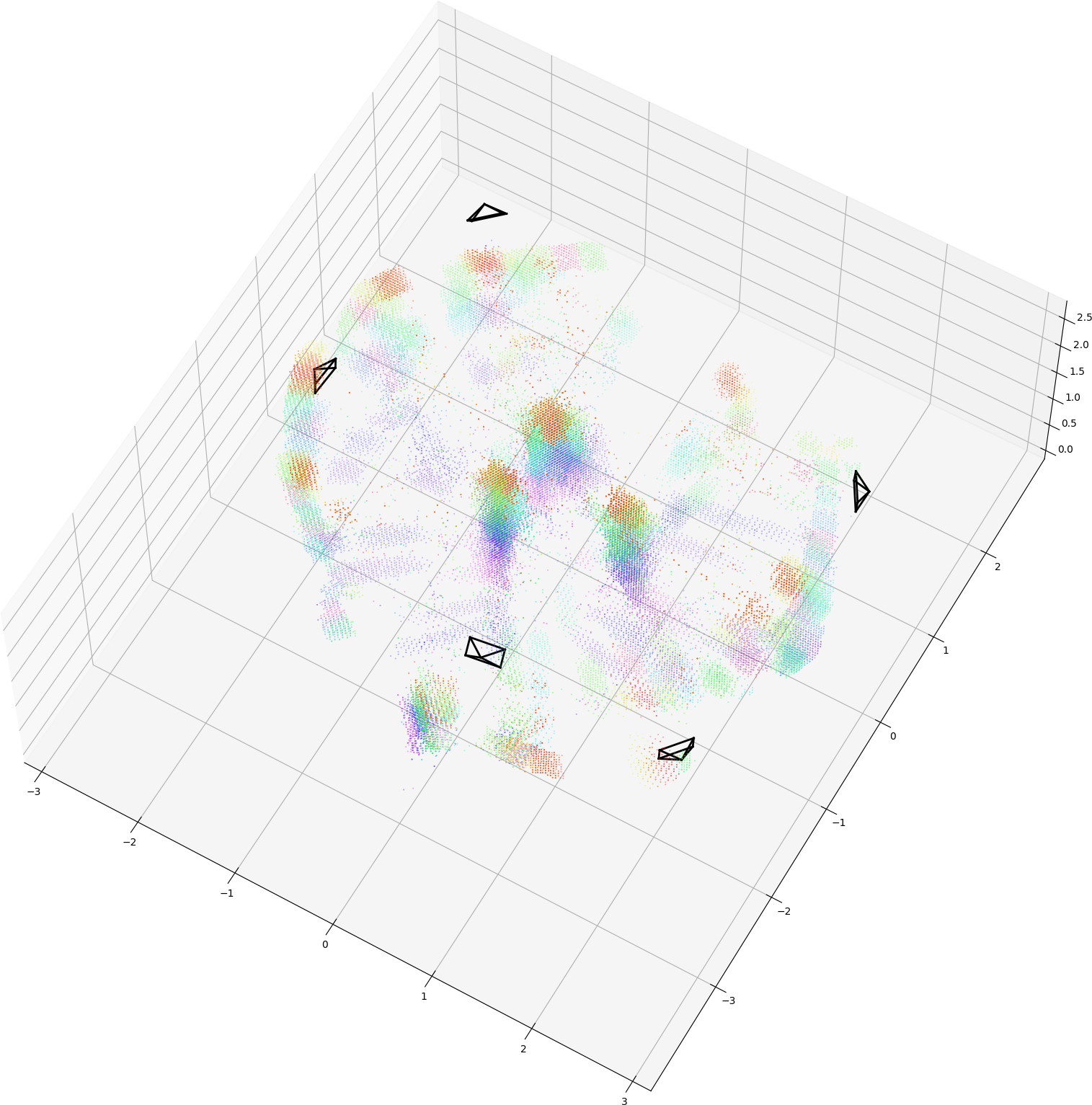}
  \end{subfigure}
  \caption{Example of depth masking in \textit{VoxelKeypointFusion}. In the left image, the keypoint heatmap beam projection of four persons from five camera views into a voxelized room is shown, with one color for each joint type. This is the default input for the peak proposal calculation. The center shows the voxelized depth images, in which three persons (the fourth is walking through the entrance) and the room's wall are clearly visible. On the right side, the projection was masked with the depth voxels. The three persons in the center of the room are now clearly visible in the projection space as well, and no peaks can be proposed between the persons anymore. Parts of the sphere's wall are still left, but since the keypoint projections do not overlap there, no peak proposals are generated from those.}
  \label{fig:vkf_depthmask}
\end{figure*}

\subsection{Related Work II}

\textit{OpenPTrack}~\cite{munaro2014openptrack, carraro2019real}, which is commonly used in robotic applications, first estimates the 2D keypoints for each image and then uses the depth images to extract the corresponding distance of each joint. For each view, a 3D person proposal is generated which is then transformed into global world coordinates. Then all proposals are assigned to a person and a joint filtering and temporal smoothing step using a \textit{Kalman-Filter} is applied. This approach was only working on the \textit{Nvidia-1080Ti} again. 
\textit{MVDeep3DPS}~\cite{kadkhodamohammadi2017multi} uses a learnable filter to drop invalid person proposals before merging them together in 3D space and refines them afterward using a calculated body-part confidence. 
\textit{Ryselis et al.}~\cite{ryselis2020multiple} followed the simple strategy of just averaging the 3D poses from the different views.
\textit{Hansen et al.}~\cite{hansen2019fusing} predicted keypoint heatmaps from the depth images and used a point cloud to estimate the center of each person before projecting the keypoint heatmaps and the depth information into a voxelized space to generate a 3D pose using a \textit{V2V}~\cite{moon2018v2v} network architecture similar to \textit{VoxelPose}. Their code is not available. 
\textit{PointVoxel}~\cite{pan2023pointvoxel} is a very recent work that follows a very similar concept, but instead of directly merging keypoint and depth voxel-maps, it has two \textit{V2V}-branches and merges the results afterward. It also includes a synthetic data generator for cross-setup generalization. Its code was not published at the time of writing.

\subsection{VoxelKeypointFusion}

For \textit{VoxelKeypointFusion} a simple depth-based masking concept was implemented. 
After generating a point cloud from the depth images, all points that fall into one voxel are counted, and if there are more than a small threshold (to drop noisy points), this voxel is considered filled. The projection space is then masked with the filled depth voxels, to drop all projection voxels that do not match to a real object. This especially helps to remove any voxel-peaks that are generated from 2D detections that do not show the same person but whose heatmap projections still overlap somewhere in 3D space.
An example of the masking can be seen in Figure~\ref{fig:vkf_depthmask}.
The depth views do not have to be aligned to the color views, but can be recorded from any direction and also come from a different number of sensors.


\section{Dataset Generalization - Part II}
\label{sec:ndgen2}
\vspace{-3pt}

\begin{table*}[htbp]
  \fontsize{7pt}{7pt}\selectfont
  \centering
  \begin{tabular}{@{}|l|c|cc|c|cc|c|c|c|c|@{}}
    \toprule
    Method \hspace{44pt}                                    & {\,}PCP{\,}   & \multicolumn{2}{c|}{PCK@100/500}           & {\,}MPJPE{\,} & \multicolumn{2}{c|}{Recall@100/500}        & {\,}Invalid{\,} & {}{\,}F1{\,}{} & Training & {\,}FPS{\,}   \\
    \midrule
    MV3DReg \cite{kadkhodamohammadi2021generalizable}       & -             & \hspace{6pt} -             & -             & 176           & \hspace{6pt} -             & -             & -               & -              & -        & -             \\
    VoxelPose                                               & 28.2          & \hspace{6pt} 10.8          & 35.9          & 119           & \hspace{6pt} 19.5          & 36.8          & 15.8            & 51.2           & none     & 19.2          \\
    VoxelPose\,(synthetic)                                  & 36.3          & \hspace{6pt} 27.8          & 65.9          & 201           & \hspace{6pt} 15.2          & 72.4          & 76.7            & 35.3           & 7h       & 8.3           \\
    Faster-VoxelPose                                        & 43.3          & \hspace{6pt} 31.1          & 55.2          & 120           & \hspace{6pt} 29.1          & 56.0          & 24.6            & 64.3           & none     & 29.3          \\
    Faster-VoxelPose\,(synthetic)                           & 37.9          & \hspace{6pt} 28.1          & 45.5          & \textbf{109}  & \hspace{6pt} 29.4          & 46.1          & \textbf{7.7}    & 61.5           & 2h       & \textbf{30.0} \\
    MvP                                                     & 0             & \hspace{6pt} 0             & 0.1           & 343           & \hspace{6pt} 0             & 0.1           & 99.9            & 0.1            & none     & 8.8           \\
    PRGnet                                                  & 4.9           & \hspace{6pt} 3.6           & 6.2           & 120           & \hspace{6pt} 3.4           & 6.4           & 44.3            & 11.5           & none     & 11.9          \\
    TEMPO                                                   & 10.4          & \hspace{6pt} 7.9           & 12.6          & 102           & \hspace{6pt} 8.8           & 12.7          & 14.0            & 22.2           & none     & 20.3          \\
    SelfPose3d                                              & 48.8          & \hspace{6pt} 36.2          & 67.7          & 143           & \hspace{6pt} 31.1          & 70.2          & 36.5            & 66.7           & none     & 13.0          \\
    mvpose                                                  & 45.9          & \hspace{6pt} 32.9          & 60.2          & 127           & \hspace{6pt} 27.1          & 61.3          & 18.5            & 70.0           & none     & 0.8           \\
    mv3dpose                                                & 1.3           & \hspace{6pt} 0.7           & 2.8           & 235           & \hspace{6pt} 0.4           & 3.1           & 57.7            & 5.8            & none     & 3.0           \\
    PartAwarePose                                           & 15.3          & \hspace{6pt} 10.3          & 25.5          & 201           & \hspace{6pt} 6.3           & 27.7          & 20.5            & 41.1           & none     & 6.9           \\
    \midrule
    VoxelKeypointFusion                                     & \textbf{54.5} & \hspace{6pt} \textbf{43.9} & \textbf{75.1} & 128           & \hspace{6pt} \textbf{35.9} & \textbf{76.6} & 24.2            & \textbf{76.2}  & none     & 11.3          \\
    \midrule
    \midrule
    MVDeep3DPS \cite{kadkhodamohammadi2017multi}            & -             & \hspace{6pt} -             & -             & 213           & \hspace{6pt} -             & -             & -               & -              & -        & -             \\
    OpenPTrack                                              & 11.7          & \hspace{6pt} 9.9           & 26.8          & 323           & \hspace{6pt} 0.8           & 33.6          & 83.9            & 21.7           & none     & 1.9           \\
    \midrule
    VoxelKeypointFusion                                     & 54.0          & \hspace{6pt} \textbf{44.2} & 72.2          & 119           & \hspace{6pt} \textbf{36.3} & 73.4          & 12.9            & \textbf{79.7}  & none     & 10.9          \\
    \bottomrule
  \end{tabular}
  \caption{Transfer to \textit{mvor}~\cite{srivastav2018mvor} without and with depth}
  \label{tab:trans_mvor}
\end{table*}

In general, the number of multi-view multi-person depth datasets is quite low. One more often used in literature is \textit{Multi View Operation Room (MVOR)}~\cite{srivastav2018mvor} which records an operation room from three different viewpoints. This is a relatively complicated dataset, since there is much occlusion, and the persons all have similar clothing. Only the upper body of a person is labeled, and most, but not all persons are labeled. 

In Table~\ref{tab:trans_mvor} it can be seen that many models have great problems with this setup, and only a few of them, including \textit{VoxelKeypointFusion}, show a decent performance on this dataset.
In comparison between \textit{VoxelPose} and \textit{Faster-VoxelPose} after synthetic training, it seems that the person-proposal step of the first one is more robust to strongly occluded lower bodies, but on the other hand, tends to predict much more invalid persons.
\textit{OpenPTrack} was quite slow, which is partially caused by the processing of multiple views on a single device and partially because of a motion lag due to the joint smoothing. To reduce the impact in terms of \textit{MPJPE}, the input frame rate was reduced in a way that the joints became relatively stable before extracting the locations. 
\textit{Hansen et al.}~\cite{hansen2019fusing} were using 4-fold cross-validation, training on three parts of the dataset, while all other models did not see the dataset before, and also used ground-truth bounding boxes for each person, which is why their score is not directly comparable.
The main improvement of the depth masking of \textit{VoxelKeypointFusion} is the reduction of the number of invalid persons, since now only persons can be predicted where the depth volume is filled. A reason why this also leads to some missing joints is, that if the keypoint heatmaps are not accurate and create a peak slightly next to a person, it gets cut off.

\begin{table*}[htbp]
  \vspace{3pt}
  \fontsize{7pt}{7pt}\selectfont
  \centering
  \begin{tabular}{@{}|l|c|cc|c|cc|c|c|c|c|@{}}
    \toprule
    Method \hspace{56pt} & {\,}PCP{\,}   & \multicolumn{2}{c|}{PCK@100/500}           & {\,}MPJPE{\,} & \multicolumn{2}{c|}{Recall@100/500}       & {\,}Invalid{\,} & {}{\,}F1{\,}{} & Training & {\,}FPS{\,}  \\
    \midrule
    VoxelPose            & 98.5          & \hspace{6pt} 97.9          & 98.7          & 19.3          & \hspace{6pt} 98.7          & 98.7          & 1.1             & 98.8           & -        & 8.0           \\
    Faster-VoxelPose     & 99.4          & \hspace{6pt} 98.6          & \textbf{99.9} & 20.5          & \hspace{6pt} 99.7          & \textbf{99.9} & \textbf{1.0}    & \textbf{99.5}  & -        & \textbf{18.0} \\
    MvP                  & 97.6          & \hspace{6pt} 97.2          & 98.3          & 18.7          & \hspace{6pt} 98.0          & 98.5          & 15.8            & 90.8           & -        & 8.9           \\
    PRGnet               & \textbf{99.5} & \hspace{6pt} \textbf{99.1} & \textbf{99.9} & 17.1          & \hspace{6pt} \textbf{99.9} & \textbf{99.9} & 2.0             & 99.0           & -        & 6.8           \\
    TEMPO                & 98.1          & \hspace{6pt} 97.4          & 98.5          & \textbf{16.8} & \hspace{6pt} 98.4          & 98.4          & 2.4             & 98.0           & -        & 5.1           \\
    SelfPose3d           & 99.3          & \hspace{6pt} 98.7          & 99.8          & 24.9          & \hspace{6pt} 99.7          & \textbf{99.9} & 8.0             & 95.7           & -        & 7.1           \\
    \midrule
    \midrule
    mvpose               & 90.5          & \hspace{6pt} 75.9          & 97.5          & 83.6          & \hspace{6pt} 73.5          & 98.5          & 10.0            & 94.0           & none     & 0.1           \\
    mv3dpose             & 84.5          & \hspace{6pt} 79.4          & 86.1          & 48.8          & \hspace{6pt} 81.8          & 86.4          & 15.6            & 85.4           & none     & 1.3           \\
    PartAwarePose        & 89.8          & \hspace{6pt} 79.9          & 92.1          & 60.5          & \hspace{6pt} 83.1          & 93.0          & 1.4             & 95.8           & none     & 1.5           \\
    VoxelKeypointFusion  & \textbf{97.1} & \hspace{6pt} \textbf{94.0} & \textbf{99.7} & \textbf{47.8} & \hspace{6pt} \textbf{97.3} & \textbf{99.9} & 2.4             & 98.7           & none     & 4.2           \\
    \midrule
    OpenPTrack           & 83.0          & \hspace{6pt} 70.9          & 95.1          & 97.6          & \hspace{6pt} 68.9          & 97.2          & 15.5            & 90.4           & none     & 1.8           \\
    VoxelKeypointFusion  & 92.6          & \hspace{6pt} 90.0          & 96.9          & 60.1          & \hspace{6pt} 85.4          & 97.8          & \textbf{0.1}    & \textbf{98.9}  & none     & 4.0           \\
    \bottomrule
  \end{tabular}
  \caption{Replication of \textit{panoptic} results and transfer without and with depth}
  \label{tab:res_panoptic}
\end{table*}

\begin{table*}[!hbp]
  \fontsize{7pt}{7pt}\selectfont
  \centering
  \begin{tabular}{@{}|l|c|cc|c|cc|c|c|c|@{}}
    \toprule

  Method \hspace{44pt} & {\,}PCP{\,}   & \multicolumn{2}{c|}{PCK@100/500}           & {\,}MPJPE{\,} & \multicolumn{2}{c|}{Recall@100/500}       & {\,}Invalid{\,} & {}{\,}F1{\,}{} & {\,}FPS{\,}   \\
  \midrule
  VoxelPose & 37.6 & 26.6 & 43.9 & 151 & 28.9 & 44.7 & 35.9 & 45.3 & 13.9 \\
  VoxelPose\,(synthetic) & 73.3 & 57.4 & 86.1 & 118 & 50.8 & 88.3 & 38.5 & 69.8 & 10.9 \\
  Faster-VoxelPose & 72.1 & 50.0 & 85.3 & 118 & 47.9 & 87.2 & 48.2 & 60.9 & 26.2 \\
  Faster-VoxelPose\,(synthetic) & 73.5 & 58.2 & 79.7 & 90.0 & 60.8 & 80.2 & \textbf{14.7} & 78.4 & \textbf{27.2} \\
  MvP & 1.8 & 1.1 & 4.2 & 385 & 0.0 & 5.6 & 97.6 & 3.2 & 8.9 \\
  PRGnet & 47.1 & 35.7 & 52.9 & 165 & 32.4 & 54.0 & 38.0 & 47.6 & 12.1 \\
  TEMPO & 59.1 & 48.2 & 63.0 & 87.8 & 48.9 & 63.2 & 29.7 & 58.0 & 11.4 \\
  SelfPose3d & 60.3 & 44.7 & 72.0 & 132 & 41.3 & 73.5 & 69.6 & 39.7 & 8.1 \\
  mvpose & 77.6 & 64.1 & 85.2 & 86.0 & 68.5 & 85.7 & 28.1 & 75.7 & 0.4 \\
  mv3dpose & 59.6 & 51.1 & 63.5 & 131 & 51.2 & 64.4 & 28.0 & 61.3 & 2.5 \\
  PartAwarePose & 73.9 & 65.8 & 79.0 & 97.8 & 70.4 & 79.7 & 22.8 & 73.2 & 4.8 \\
  VoxelKeypointFusion & \textbf{85.3} & \textbf{73.4} & \textbf{93.8} & \textbf{82.0} & \textbf{77.4} & \textbf{94.2} & 27.2 & \textbf{80.5} & 8.4 \\
    \bottomrule
  \end{tabular}
  \caption{Averaged generalization on all four unseen datasets (\textit{human36m}, \textit{shelf}, \textit{campus}, \textit{mvor}).}
  \label{tab:trans_avg}
\end{table*}

\vspace{3pt}
At last the \textit{Panoptic}~\cite{joo2015panoptic} dataset is evaluated, and the results can be found in Table~\ref{tab:res_panoptic}. The cameras are mounted in a dome-like structure and point to the center of it. Most learned algorithms are trained using this dataset, and their results could be closely replicated (\textit{Faster-VoxelPose} published a different backbone model than what was described in its paper). The evaluated cameras are the same as in the previous works~\cite{voxelpose,fastervoxelpose,wang2021mvp,lin2021multi,wu2021graph}, but the frames were changed to contain consecutive motions, as described before. To make the results of all models comparable, only three out of four test scenes were used, since for \textit{160906\_band4} no depth data was available. Note that since the depth cameras were not time synchronized, their alignment to the color images and to the pose labels is not perfect. They were considered as belonging together if the time difference was below a threshold.

From the RGB-based algorithmic approaches \textit{VoxelKeypointFusion} performs best, and is only close behind the models that were trained on this dataset in terms of detection reliability. 
Some of the invalid predictions might come from persons entering the room, which often are unlabeled.

\vspace{3pt}
The general goal of all developed pose estimators is their usage in practical applications. Since it is not a practical option to create a training dataset matching the exact setup of the application, because every time someone would need to create and label a new dataset, the generalization to unseen datasets is a crucial aspect. This is also important since the models have to be robust to all kinds of situations, and not just to the ones they were trained on. As could be seen in the experiments above, all previous models have setups in which they show good, and others in which they show bad performance. For a new setup without labels it is hard to predict which it will be. \textit{VoxelKeypointFusion} on the other hand shows a good performance on all, even if it might not always be the best. This can be easily seen in Table~\ref{tab:trans_avg}, which presents the average performance across all unseen datasets.


\section{Ablation studies}
\label{sec:ablations}

One limitation of \textit{VoxelKeypointFusion} is, that since there is no neural refinement, the location accuracy is highly dependent on the voxel-resolution. As can be seen in Table~\ref{tab:abls}, a lower voxel-resolution leads to better results, even though it does not directly scale (but $25mm$ is even better than the result with $50mm$ in Table~\ref{tab:trans_h36m}). The voxel size also has a significant impact on the inference speed. As expected, with more cameras the results get better, but the inference time increases as well. With fewer cameras the performance drops, and the model predicts more invalid persons because overlaps where no persons stand are not pruned that much by different viewpoints that would see an empty space there. The number of persons, especially if there are many occlusions, has a notable impact on the inference speed as well. The total 3D time for \textit{Shelf} is around $48ms$, for \textit{Campus} around $56ms$, and for \textit{Panoptic}, with fewer voxels but more total and occluded keypoints than \textit{Campus}, around $100ms$.

\begin{table*}[htbp]
  \fontsize{7pt}{7pt}\selectfont
  \centering
  \begin{tabular}{@{}|l|c|cc|c|cc|c|c|c|@{}}
    \toprule
    Method \hspace{44pt} & {\,}PCP{\,}   & \multicolumn{2}{c|}{PCK@100/500} & {\,}MPJPE{\,} & \multicolumn{2}{c|}{Recall@100/500}       & {\,}Invalid{\,} & {}{\,}F1{\,}{} & {\,}FPS{\,}  \\
    \midrule
    VKF\,(h36m,\,voxelsize=25)        & 97.1 & \hspace{6pt} 82.6 & 100  & 62.1 & \hspace{6pt} 95.0 & 100  & 0    & 100  & 3.4  \\
    VKF\,(h36m,\,voxelsize=100)       & 95.1 & \hspace{6pt} 71.2 & 100  & 76.5 & \hspace{6pt} 94.5 & 100  & 0    & 100  & 9.7  \\
    VKF\,(panoptic,\,cameras=3)       & 89.1 & \hspace{6pt} 81.4 & 95.3 & 77.7 & \hspace{6pt} 78.6 & 96.9 & 10.8 & 92.9 & 6.8  \\
    VKF\,(panoptic,\,cameras=10)      & 98.5 & \hspace{6pt} 94.7 & 99.7 & 43.3 & \hspace{6pt} 98.9 & 99.8 & 3.9  & 97.9 & 2.2  \\
    \bottomrule
  \end{tabular}
  \caption{Ablation experiments with \textit{VoxelKeypointFusion}.}
  \label{tab:abls}
\end{table*}


\section{Whole-body estimation}
\label{sec:wholebody}
\vspace{-3pt}

One significant benefit of algorithmic approaches over the learning-based ones, besides a simpler transfer to other setups, is their ability to handle different input keypoints, because they do not require a (different) training dataset including exactly those keypoints. Instead of predicting 3D joint locations of humans, they also work for animals, or following the idea of \textit{Pose for Everything}~\cite{xu2022pose}, for any object one requires. Another option is their extension to predict whole-body keypoints, which include additional face, foot and finger keypoints.

\begin{table*}[htbp]
  \vspace{3pt}
  \fontsize{7pt}{7pt}\selectfont
  \centering
  \begin{tabular}{@{}|l|c|cc|c|cc|c|c|c|@{}}
    \toprule
    Method \hspace{2pt}  & {\,}PCP{\,} & \multicolumn{2}{c|}{PCK@100/500} & {\,}MPJPE (All|Body|Face|Hands){\,} & \multicolumn{2}{c|}{Recall@100/500}       & {\,}Invalid{\,} & {\,}FPS{\,} \\
    \midrule
    VKF                  & 99.7        & \hspace{6pt} 96.8 & 99.8         & 40.5 | 37.4 | 38.3 | 40.7           & \hspace{6pt} 98.5 & 100                   & 0.5             & 4.9         \\
    \bottomrule
  \end{tabular}
  \caption{Whole-body estimation with \textit{VoxelKeypointFusion} on \textit{h3wb}.}
  \label{tab:h3wb}
\end{table*}

To evaluate the performance, the \textit{h3wb} dataset~\cite{Zhu_2023_ICCV} was used, which extended a part of the \textit{human36m} dataset with whole-body keypoints (17 body, 6 foot, 68 face, 42 hand). For simplification of the labeling process, only frames with good visibility of the actor were used, therefore the dataset is somewhat simpler than the original one (the default model of \textit{VoxelKeypointFusion} reaches a \textit{MPJPE} of $30.4\,mm$ there, and the authors estimated around $17\,mm$ labeling error). Since the test-split is not available, every $100th$ frame of the training-split was used for evaluation. The results can be found in Table~\ref{tab:h3wb}. Since this dataset is normally used for single-view 2D-to-3D pose lifting, no other results were found for comparison.

In general, the whole-body prediction works quite well, though there still are some improvement possibilities. \textit{VoxelKeypointFusion} shows some artifacts of the voxelization process, which due to the discretization pushes some keypoints closer together than they really are.
An example prediction can be seen in Figure~\ref{fig:init_example}.


\section{Conclusion}
\label{sec:conclusion}
\vspace{-3pt}

This paper showed, through an evaluation of different datasets, that the generalization of learned algorithms directly to new setups is often poor. Synthetic training for the new camera setup is a good option to achieve competitive results, but it does not help in every case. In general, all algorithms have some strengths and weaknesses, which are best noticeable if they are evaluated on different datasets.

\vspace{3pt}
With \textit{VoxelKeypointFusion} a new voxel-based algorithmic approach was presented, that showed the best generalization performance between multiple datasets while having decent speed at the same time. It also has the option to use depth data for improved results, especially reducing the number of invalid persons. This depth-masking concept might also be useful for other voxel-based approaches.

\textit{VoxelKeypointFusion} additionally shows that all the necessary person information is already present in the 2D heatmaps, which currently is ignored by the learnable voxel-based approaches. One limitation though is that since there is no neural refinement, the location accuracy is highly dependent on the voxel-resolution.

\vspace{3pt}
A significant benefit of algorithmic approaches is their generalization to other keypoints, which is especially useful for whole-body pose estimation.
\textit{VoxelKeypointFusion} was extended to predict them, making it the first multi-view multi-person whole-body pose estimation algorithm. 


\newpage

{
    \small
    \bibliographystyle{ieeenat_fullname}
    \bibliography{main}
}

\end{document}